\documentclass[lettersize,journal]{IEEEtran}
\usepackage{amsmath,amsfonts}
\usepackage{hyperref}
\usepackage{xcolor}
\usepackage{algorithmic}
\usepackage{algorithm}
\usepackage{array}
\usepackage[caption=false,font=normalsize,labelfont=sf,textfont=sf]{subfig}
\usepackage{textcomp}
\usepackage{stfloats}
\usepackage{url}
\usepackage{verbatim}
\usepackage{graphicx}
\usepackage{cite}
\usepackage[english]{babel}
\usepackage{graphicx}
\usepackage{dcolumn}
\usepackage{bm}

\begin{document}

\title{Identifying internal patterns in (1+1)-dimensional directed percolation using neural networks}

\author{Danil Parkhomenko\IEEEauthorrefmark{1}\IEEEauthorrefmark{3} \thanks{Corresponding author: Danil Parkhomenko, parkhomenko.dale@dvfu.ru},
Pavel Ovchinnikov\IEEEauthorrefmark{1}\IEEEauthorrefmark{2}\IEEEauthorrefmark{3},
Konstantin Soldatov\IEEEauthorrefmark{1}\IEEEauthorrefmark{2}\IEEEauthorrefmark{3},
Vitalii Kapitan\IEEEauthorrefmark{4},
Gennady Y.~Chitov\IEEEauthorrefmark{3}%

\thanks{\IEEEauthorrefmark{1}Institute of High Technologies and Advanced Materials, Far Eastern Federal University, Russky Island, Ajax 10, Vladivostok 690922, Russia.}%
\thanks{\IEEEauthorrefmark{2}Institute of Applied Mathematics, Far Eastern Branch, Russian Academy of Science, Radio 7, Vladivostok 690041, Russia.}%
\thanks{\IEEEauthorrefmark{3}Bogoliubov Laboratory of Theoretical Physics, Joint Institute for Nuclear Research, Dubna 141980, Russia.}%
\thanks{\IEEEauthorrefmark{4} Institute for Functional Intelligent Materials, National University of Singapore, 21 Lower Kent Ridge Road, 119077, Singapore.}%
}



\maketitle

\begin{abstract}
In this paper we present a neural network-based method for the automatic detection of phase transitions and classification of hidden percolation patterns in a (1+1)-dimensional replication process. The proposed network model is based on the combination of CNN, TCN and GRU networks, which are trained directly on raw configurations without any manual feature extraction. The network reproduces the phase diagram and assigns phase labels to configurations. It shows that deep architectures are capable of extracting hierarchical structures from the raw data of numerical experiments.
\end{abstract}

\begin{IEEEkeywords}
Directed percolation, phase transitions, deep learning, temporal convolutional networks, GRU.
\end{IEEEkeywords}

\section{Introduction}\label{sec:intro}
The kinetics and phase transitions in non-equilibrium systems have been very actively studied during the last several decades. The applications of those systems range from statistical physics to biology, ecology or network theory \cite{Hinrichsen:2000,Odor:2004,Hinrichsen:2006,Grassberger:2015,Barabasi:2016,Wetterich:2021,Stepinski:2023}. The central problem of these studies is the transition between the active and inactive (absorbing) states. The active states of such models are the phases of directed percolation (DP)  from the viewpoint of percolation theory \cite{Stauffer:1992}, and the transition into such phase belongs to the DP universality class \cite{Hinrichsen:2000,Odor:2004}. 

A progress in our understanding of percolation has been made, when the hidden hierarchical structure of percolating phase was revealed. It has been shown  \cite{Chitov:2015,Chitov:2016, ovchinnikov2025hidden,Ovchinnikov:2025}  that the active phases of several $(1+1)$ kinetic processes and some other 2D models of percolation,
possess hidden orders characterized by distinct percolative patterns, emerging at specific points of the DP phase transitions.

In the present paper we study a particular replication process with parallel update (probabilistic cellular automaton, PCA) in $1+1$ (space-time) dimensions \cite{Chitov:2015}.
Along with the trivial absorbing state, it has several phases with different percolative patterns in the active state.  This model has been studied using analytical and standard numerical (Monte Carlo, finite-size scaling) methods \cite{Chitov:2015,ovchinnikov2025hidden, Ovchinnikov:2025}. The goal of this work is to independently probe results of the direct numerical simulations \cite{ovchinnikov2025hidden, Ovchinnikov:2025} and to advance methods using neural networks to deal with percolation patterns.

Machine learning (ML) methods have opened up new avenues for research in statistical physics, including DP \cite{carrasquilla2017machine, Cheng:2021, korol2022calculation, andriushchenko2022new,  fan2023searching, kapitan2023application, shiina2024super, bayo2025machine, druz2025phase}.
ML methods are used to study critical behavior in DP, a well-known class of nonequilibrium phase transitions. Various ML algorithms predict critical points, critical exponents, and characteristic time in DP models. Siamese Neural Networks (SNN), a semi-supervised approach, predict critical values and exponents by comparing configuration data pairs for similarity \cite{shen2025learning}. Supervised learning, employing Convolutional Neural Networks (CNN) and Fully Connected Networks (FCN), classifies DP phases and determines critical properties \cite{saif2023determination}. Unsupervised learning, through autoencoders and principal component analysis can represent the capacity of connected cluster, which is the percolation order parameter \cite{shen2024machine}. Transfer learning methods like domain adversarial neural networks also effectively identify DP critical points and exponents with limited labeled data \cite{shen2022transfer}.


Building on these developments, we extend the application of machine learning to the analysis of percolation patterns in our replication PCA model. While previous studies have largely focused on detecting phase boundaries or estimating critical properties, our approach targets the finer task of identifying distinct percolative structures within the active phase. We developed a neural network that can reliably detect boundaries of distinct patterns within the percolation process. The model combines a point-wise convolution, a temporal convolutional encoder, and a GRU summarizer, followed by a multi-head classifier (see Sec. \ref{sec:model} for details). It is trained on a set of about \(1.5\times10^5\) unique configurations (10–15 epochs over the full dataset without a sampler), with system sizes \(N\in[50,100]\) and \(T\in[500,5000]\).

\paragraph*{Replication model and percolation background}
In this paper we study the replication process with spreading of active sites following stochastic update rules governed by two probability parameters $p,q \in [0,1]$. 
\begin{figure}[!t]
    \centering
    \includegraphics[width=\linewidth]{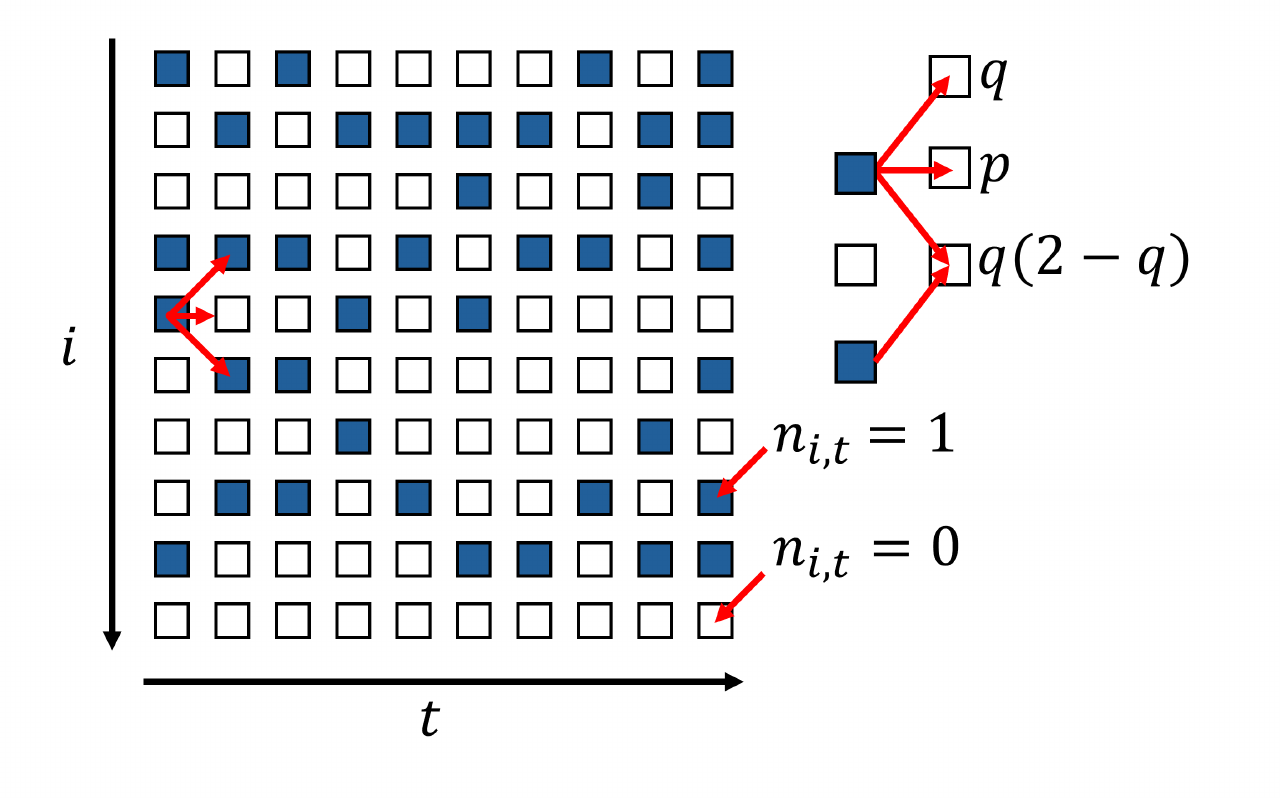}
    \caption{Space--time lattice of binary states with axes \(i\) (vertical) and \(t\) (horizontal); filled blue cells denote \(n_{i,t}=1\), empty cells \(n_{i,t}=0\).
    Local stochastic update rules: survival of the center with probability \(p\), activation from each neighbor with probability \(q\); two active neighbors yield \(q(2-q)\).}
    \label{fig:pca}
\end{figure}

The state of the system at discrete time step $t = 0,1, \ldots, T$ is specified by the occupation numbers $n_{i,t} = 0,1$ (empty/filled), $i = 1,\ldots, N$, with periodic boundary conditions (PBC) in the spatial direction, see Fig.~\ref{fig:pca}.  At $t=0$ the initial random configuration of $n$-s is cast.
The probabilistic rules presented in Table~\ref{RulesTab} and visualized in Fig.~\ref{fig:pca}
are used to numerically mimic stochastic temporal evolution of the model. 
This PCA becomes the BDP model in the limit $p \to 0$ and the pair contact process in the limit  $q \to 0$ \cite{Hinrichsen:2000}.


\begin{table}[!t]
\centering
\caption{Stochastic update rules} 
\label{RulesTab}
\setlength{\tabcolsep}{4.5pt}\renewcommand{\arraystretch}{1.20}
\begin{tabular}{|l|llll|ll|l|l|}
\hline
\begin{tabular}[c]{@{}l@{}}$n_{i-1,t}$\\ $n_{i,t}$\\ $n_{i+1,t}$\end{tabular}                   & \multicolumn{1}{l|}{\begin{tabular}[c]{@{}l@{}}1\\ 1\\ 1\end{tabular}} & \multicolumn{1}{l|}{\begin{tabular}[c]{@{}l@{}}1\\ 1\\ 0\end{tabular}} & \multicolumn{1}{l|}{\begin{tabular}[c]{@{}l@{}}0\\ 1\\ 1\end{tabular}} & \begin{tabular}[c]{@{}l@{}}0\\ 1\\ 0\end{tabular} & \multicolumn{1}{l|}{\begin{tabular}[c]{@{}l@{}}0\\ 0\\ 1\end{tabular}} & \begin{tabular}[c]{@{}l@{}}1\\ 0\\ 0\end{tabular} & \multicolumn{1}{c|}{\begin{tabular}[c]{@{}c@{}}1\\ 0\\ 1\end{tabular}} & \begin{tabular}[c]{@{}l@{}}0\\ 0\\ 0\end{tabular} \\ \hline
\begin{tabular}[c]{@{}l@{}}Probability of\\ $n_{i,t+1} = 1$\end{tabular} & $p$                                                                      &                                                                        &                                                                        &                                                   & $q$                                                                      &                                                   & $q(2-q)$                                                  & $0$                                                  \\ \hline
\end{tabular}
\end{table}

\begin{figure}[!t]
\centering
\includegraphics[width=7.2cm]{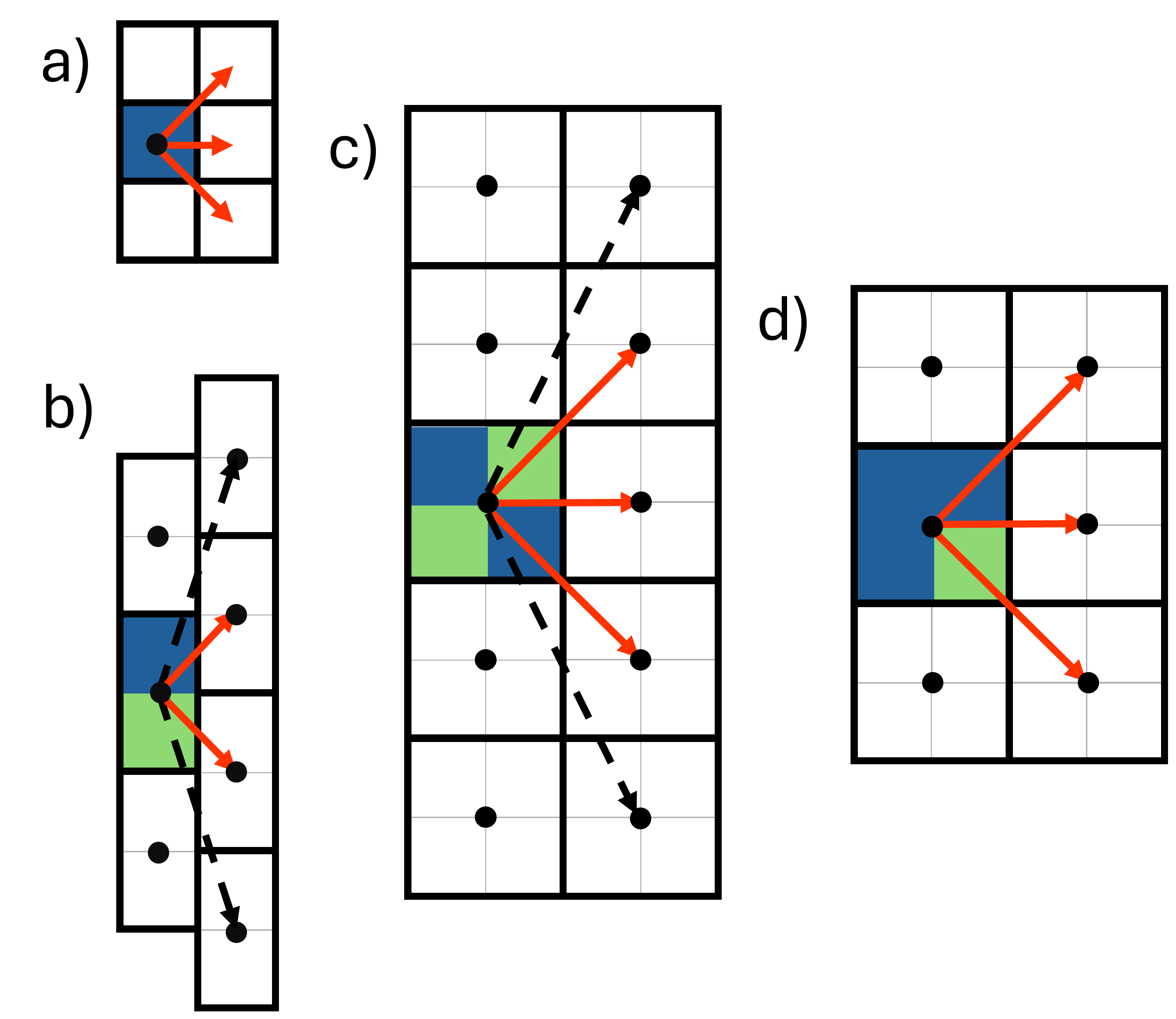}
\caption{Schematic percolation patterns considered in this work. Active (occupied) sites are shown in blue, empty sites in green. (a) Percolation on the original lattice sites. (b) Dipolar percolation: phase \(D\) accounts for nearest neighbors (solid red lines), while phase \(D^{+}\) includes also next-nearest neighbors (dashed black lines). (c) Similarly for quadrupole patterns \(Q\) and \(Q^{+}\). (d) Plaquette pattern \(PL\) includes either nearest fully occupied \(2\times2\) blocks or blocks with one empty site.}
\label{Patterns}
\end{figure}

The phase boundary between the absorbing and active (percolating) phases of the replication process is shown in Fig.~\ref{fig:pred_phase_diagr_50x2000}. Our main focus will be on the hidden percolating patterns of the active phase. 
We refer to a percolation pattern as a spanning cluster that traverses the entire system. This cluster is formed by renormalized sites obtained from local subsets of the original lattice. The assignment of active/inactive labels to renormalized sites is not unique. For instance, for \(2\times2\) plaquettes one may define the quadrupole configuration as an active renormalized site and treat all other local configurations as inactive; alternatively, one may label as active only fully occupied plaquettes or those with three occupied sites. See Fig.~\ref{Patterns} for the pattern conventions used in this work and Ref.~\cite{Ovchinnikov:2025} for more technical details.

Edges between renormalized sites need not coincide with nearest-neighbor bonds of the original lattice. This is natural from a network point of view (e.g., clusters of mutually acquainted pairs on a social map). In our model, the update rules admit percolative connections among the nearest and next-nearest renormalized neighbors.

\paragraph*{Space–time sizes and data view}
A single realization is a Boolean array \(X\in\{0,1\}^{N\times T}\) with \(N\) spatial sites and \(T\) time steps (periodic boundary conditions in space). This space–time field can be seen either as a binary image (width \(N\), height \(T\)) or as a multivariate time series of length \(T\) with \(N\) binary channels. We exploit the latter: temporal convolutions (TCN) and a GRU summarize the sequence, while a pointwise \(1{\times}1\) convolution mixes spatial channels. This representation makes the model length-agnostic in \(T\), enabling inference at \(T\neq T_{\text{train}}\).

\begin{figure}[!t]
    \centering
    \includegraphics[width=0.9\linewidth]{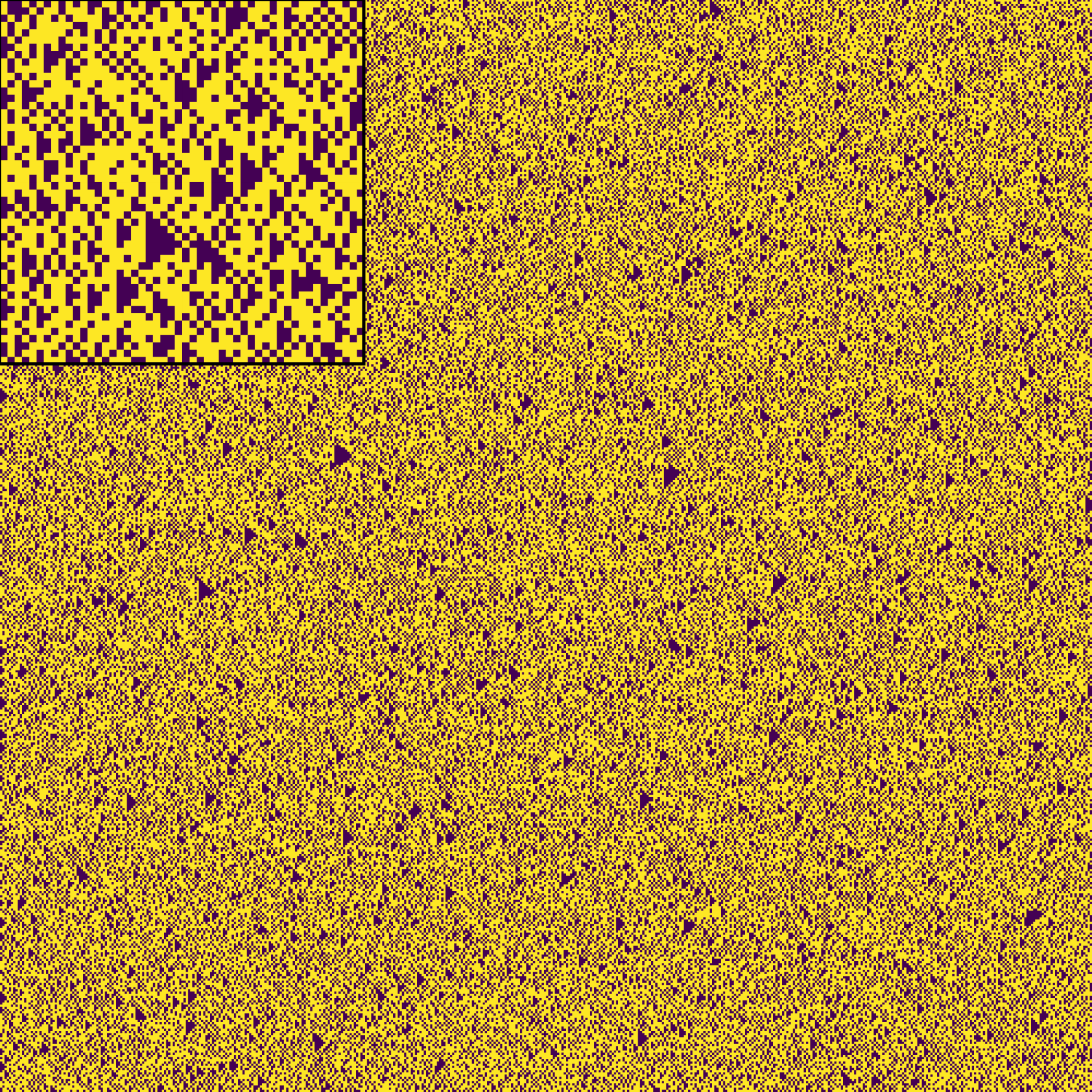}
    \caption{Zoomed view of a representative \(PL\) configuration at \(p=0.51\), \(q=0.90\).}
    \label{fig:pl-zoom}
\end{figure}

\section{Data}\label{sec:data}
The data consist of Boolean arrays; therefore, in the conventional notation for sequential data of dimensionality $(N, L, F)$, we have a batch of size $N$, a sequence length $L \equiv T$, and a feature dimension $F \equiv N$.

The target variable indicates the presence of a percolation process in the system and, if such a process is observed, identifies the specific pattern that emerges. Thus each observation corresponds to a phase transition with a specific pattern. 
To plot the phase diagram, calibration should be performed in a fixed order (The notations are explained in Fig.~\ref{Patterns})  
\[
A \;-\; PL \;-\; Q^{+} \;-\; D^{+} \;-\; Q \;-\; D .
\]
The same order is used in the target vector. For example, a percolating plaquette ($PL$) sample has
\([0,\,1,\,0,\,0,\,0,\,0]\) for train based on multi-hot targets.
The percolating state is omitted because it is logically equivalent to \(1-\)Absorbing.

To suppress stochasticity near boundaries (where different patterns may percolate at neighboring \((p,q)\)), we simulate much longer in time than in space; empirically, ratios \(N{:}T \approx 1{:}10\) stabilize phase assignment.

As a first branch we study systems with \(N=50\) and \(T=1000\). Sampling strategy impacts the training stability and the fidelity of the  phase diagram: (i) boundary-focused sampling inflates the variance of the PCA-derived labels and destabilizes training, particularly for small systems; (ii) incomplete or non-uniform coverage of the \((p,q)\) plane induces localized systematic errors; (iii) sampling deep inside each phase provides the most reliable supervision and yields maps with sharper boundaries and improved spatial coherence.

\paragraph*{On-the-fly loading}
Keeping all Boolean arrays in RAM is impractical. Each \((N,T)\) system is stored as a serialized compressed bitstream together with two numbers: the byte offset of its start in the file and the byte length to read in order to reconstruct that sample (system + target).%

\paragraph*{System generation and (p,q) coverage}
We generate datasets in two modes: (i) at a fixed set of “special” points well inside each phase, and (ii) at uniformly random points used as a near out-of-sample stress test.

As has been shown in Sec.~\ref{sec:data}, we select several $(p,q)$ points deep in the interior of each phase and generate $4096$ independent systems per point (different random initial states and RNG seeds). Empirically, this cardinality is sufficient to capture the local stochastic variability of the dynamics without incurring diminishing returns in accuracy; increasing beyond $\sim\!4\,$k configurations per point produces negligible metric gains while linearly increasing the training set size and computing costs. In total, this yields about $1.56\times 10^5$ systems and ensures a stable training process.

Additionally, we draw $100$ points uniformly at random in $[0,1]\times[0,1]$ over $(p,q)$ and generate $128$ systems per point (about $1.28\times 10^4$ systems in total). The probability of coinciding with a special point is negligible, so this set serves as a “nearly out-of-sample” distribution to probe robustness to distributional shift and unseen parameter combinations. On the other hand, we generate a set of $10^5$ systems with random $(p, q)$ combinations, which is used for testing and evaluation.

The main series uses $N\in\{50,100\}$ and $T\in[500,5000]$. When constructing phase maps, we adopt $N{:}T\approx 1{:}10$, which suppresses boundary fluctuations and stabilizes label assignment (Sec.~\ref{sec:calibration}).

For the probability-sweep experiments at fixed $q$ (Sec.~\ref{sec:critical}), we generate dense grids over $p$ and average statistics over $1024$ independent realizations per grid point.

\section{Network Model}\label{sec:model}
The performance of the network model is evaluated against deterministic labels obtained by the cellular-automation–based connectivity algorithm (PCA-based pipeline from the previous work \cite{Ovchinnikov:2025}). Our modeling goal is approximating the pattern and percolation distributions as a multitarget task with faster and more stable neural network. We expect a slight relative deviation from the critical points reported in the previous work.

The final model is organized as follows.  
First, a point-wise \(1\times1\) convolution
\[
\operatorname{Conv}_{1\times1}(N \rightarrow d_{\mathrm{model}})
\]
linearly mixes spatial channels at each time step and projects to a convenient width.

Next, we adopt a temporal convolutional network (TCN) as the encoder (we experimented with transformer variants such as Linformer, MEGA, BERT, ViT, but a TCN proved simpler and more stable for this task). A GRU then summarizes the encoded sequence into a hidden state, which is passed to a small MLP producing six logits (one per head). 
The \emph{Percolating} head is dropped since it equals \(1-\)Absorbing. 
Training uses binary cross-entropy losses without class weighting.

\subsection{Architecture and length generalization}\label{sec:arch-length}
\begin{figure}[!t]
    \centering
    \includegraphics[width=\linewidth]{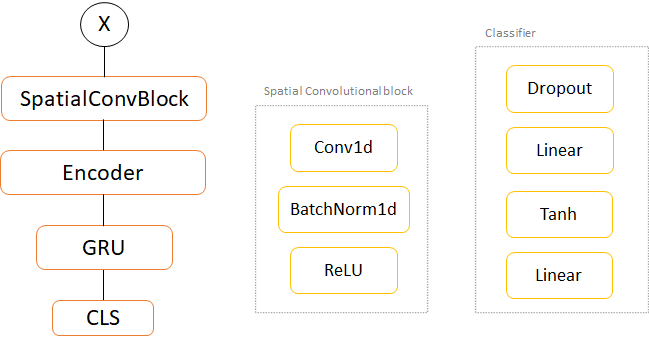}
    \caption{Base model architecture for the BDP task.}
    \label{fig:base-arch}
\end{figure}

We train a lightweight TCN--GRU on \(N=50,\,T=1000\), then evaluate at shorter/longer \(T\) without retraining; quality improves with \(T\), indicating genuine temporal learning.
\begin{figure*}[!t]
    \centering
    \subfloat[$T{=}500$]{\includegraphics[width=0.31\textwidth]{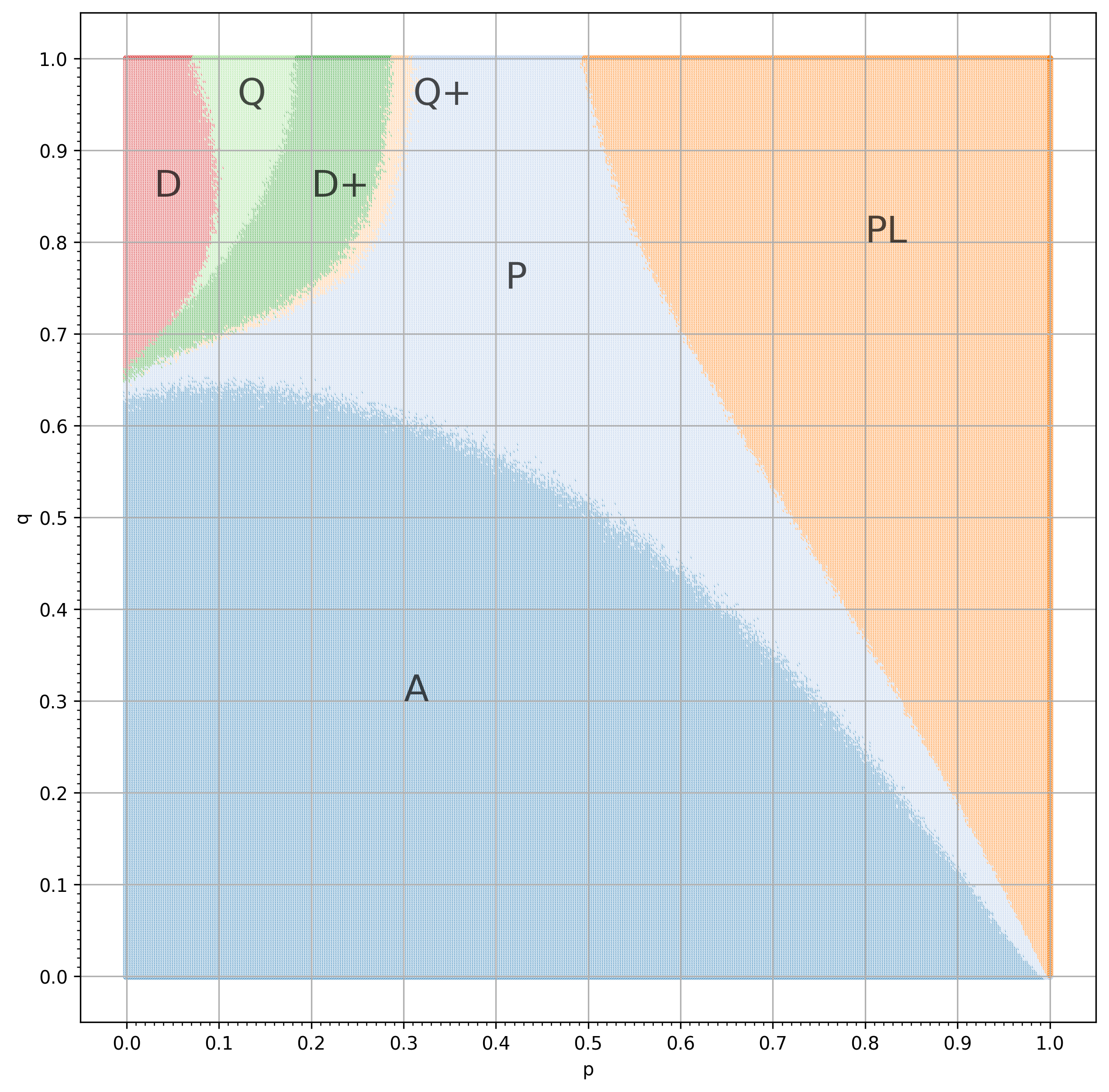}}\hfil
    \subfloat[$T{=}1000$]{\includegraphics[width=0.31\textwidth]{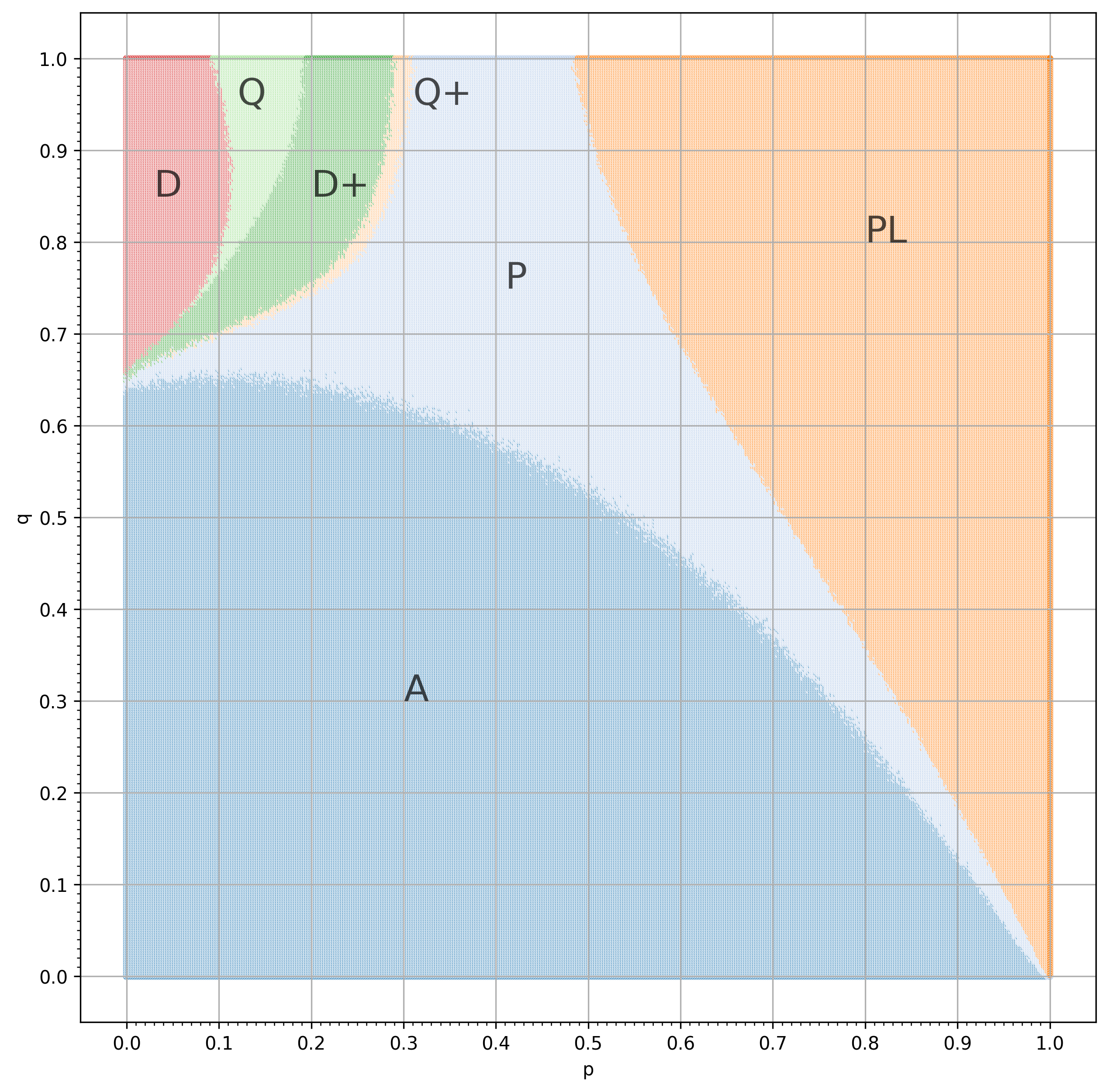}}\hfil
    \subfloat[$T{=}2000$]{\includegraphics[width=0.31\textwidth]{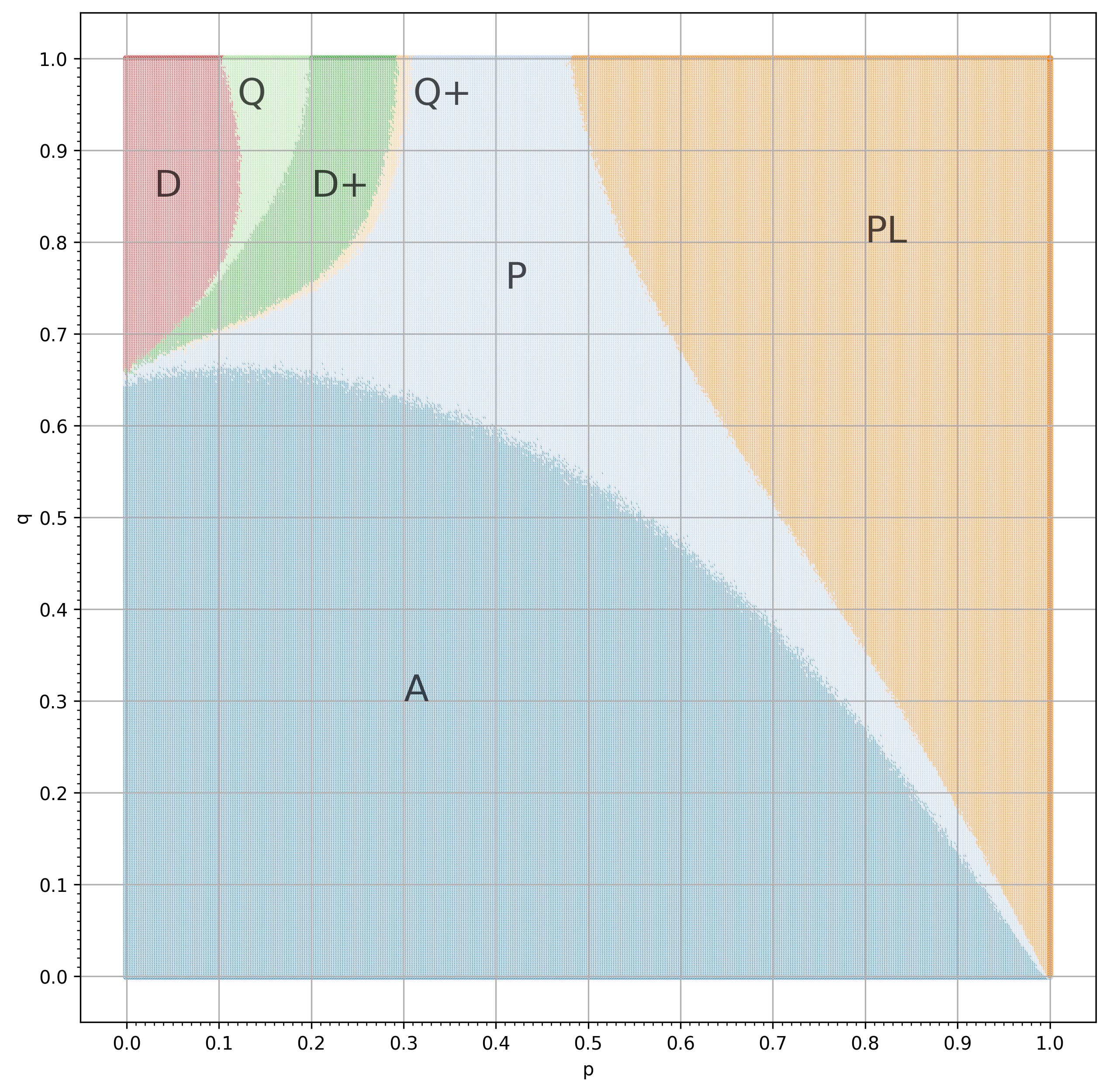}}
    \caption{One model—multiple sequence lengths. A TCN--GRU trained at $N=50$, $T=1000$ applies directly to $T/2$ and $2T$; boundaries sharpen with $T$.}
    \label{fig:phase-multiT}
\end{figure*}

The network model is multi-head (multi-target): binary classification per head with noticeable class imbalance, so logit scales differ across heads.

\subsection{Calibration}\label{sec:calibration}
The neural network outputs pointwise scores representing the learned mapping from \((N,T)\) systems to pattern-membership signals. Since head-wise score distributions differ (and are shifted due to imbalance e.g., \emph{Percolating} is frequent wherever the system is non-absorbing, while specific patterns are rarer), calibration is needed to draw a single phase map.

\paragraph*{Two calibration strategies}
\begin{enumerate}
\item \textbf{Auxiliary mapper (scores\(\to\)single class).}  
A shallow feed-forward network with SwiGLU (Swish-Gated Linear Unit) activation is trained with cross-entropy on raw scores to output a multiclass prediction, e.g.,
\([0.00,0.94,0.01,0.09,0.12,0.03]\to [0.00,0.97,0.00,0.01,0.02,0.00]\to D\).
Pros: can consume a subset of heads and reveals which heads best rank others (e.g., \(Q^{+}\) strongly ranks \(Q^{+},D^{+},Q,D\)).  
Cons: needs a separate small fit per chosen subset (fast in practice).
\item \textbf{Per-head PR thresholding.}  
For each head we sweep the precision–recall curve on validation and choose the F1-maximizing threshold.  
Pros: simple and universal; mirrors the training task.  
Cons: metrics reflects class prevalence (our \((p,q)\) sampling is geometrically balanced but label frequencies are not 0.5), which can bias thresholds; unlike (1), this cannot mix arbitrary subsets of heads.
\end{enumerate}

When visualizing probability sweeps versus \(p\) at fixed \(q\), we also show the head-wise thresholds as semi-transparent (below threshold) vs solid (above threshold) points; crossing the threshold gives an estimate of a critical point for that pattern.
\subsection{Metrics}\label{sec:metrics}
Due to class imbalance and heterogeneous “difficulty” across patterns, we report a combination of threshold-free and thresholded metrics.

For each head $k\!\in\!\{A,PL,Q^{+},D^{+},Q,D\}$ we compute ROC-AUC and PR-AUC. ROC-AUC measures the overall ranking ability of score $z_k$, whereas PR-AUC is more informative for rare positive classes and thus near phase boundaries.

For the phase diagram visualization we select per-head decision thresholds $\tau_k$ by maximizing $F_1$ on a held-out validation split (see Sec.~\ref{sec:calibration}). We report $F_1(\tau_k)$ and class-wise error rates at these operating points. Since label prevalences differ substantially, $F_1$ may be mildly biased toward majority states; therefore, we always accompany it with ROC–AUC/PR-AUC.

\begin{figure*}[!t]
    \centering
    \subfloat[Full phase diagram]{\includegraphics[width=0.45\textwidth]{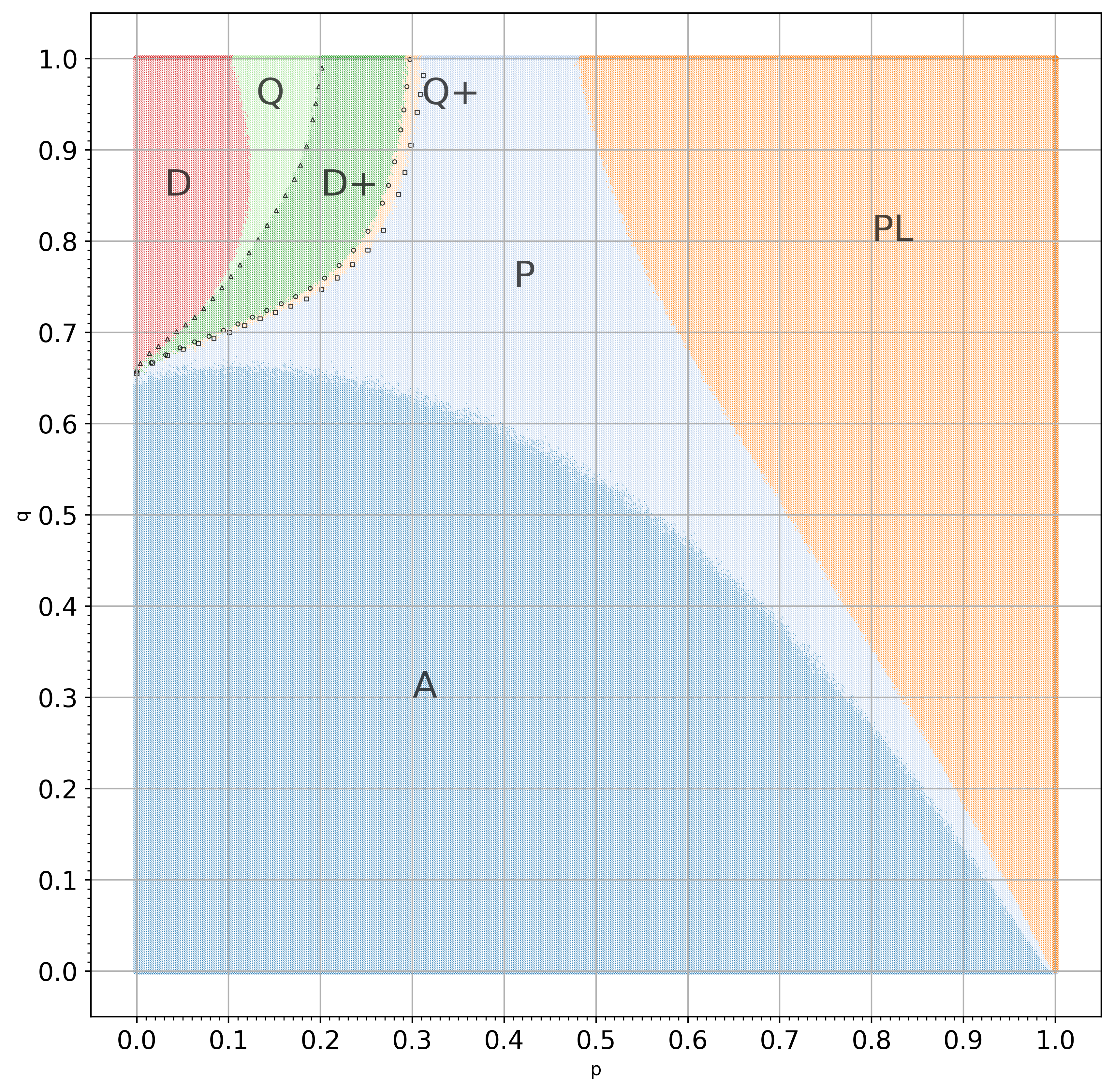}}\hfil
    \subfloat[Zoom: $p < 0.4$, $q < 0.6$]{\includegraphics[width=0.45\textwidth]{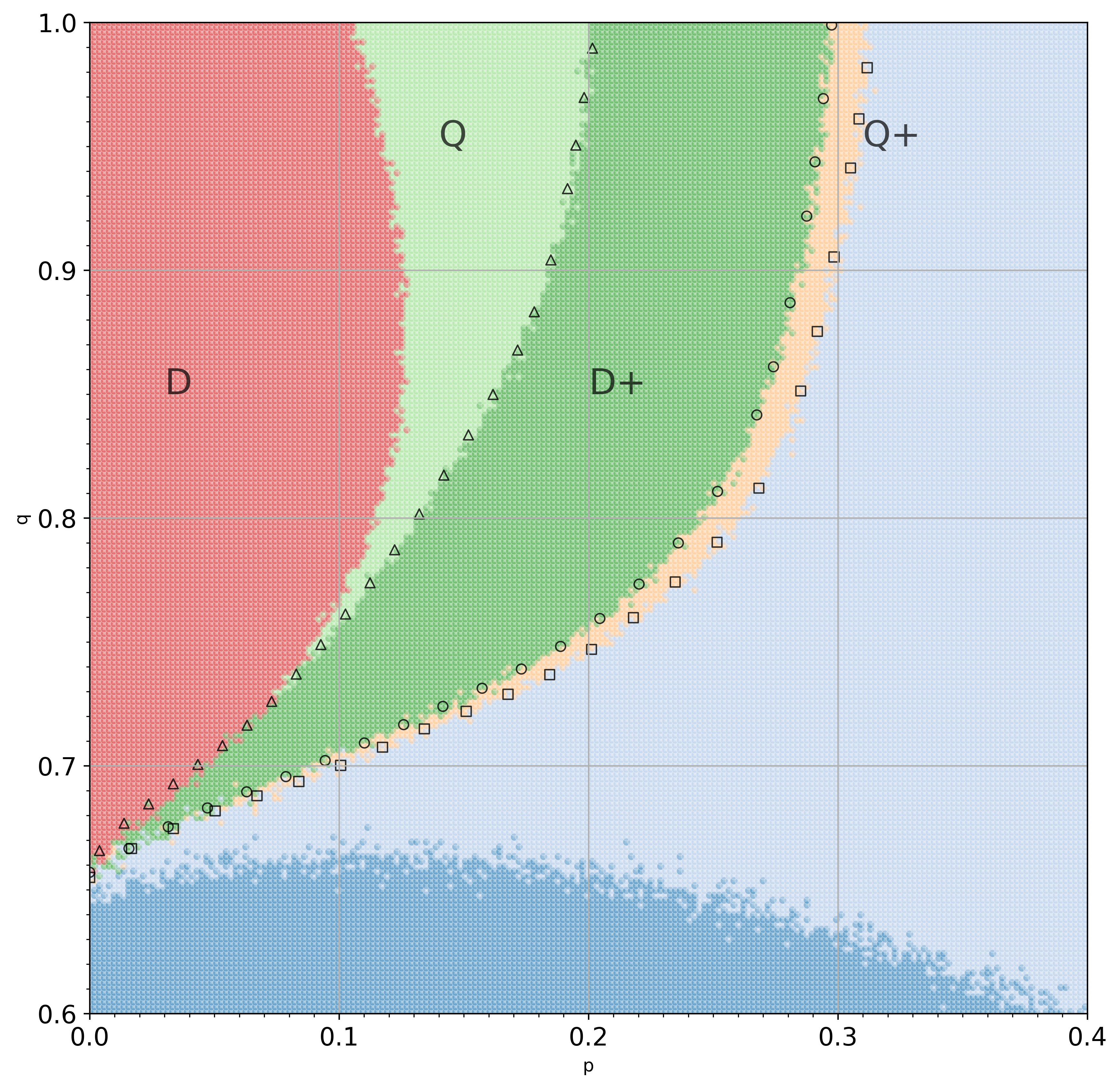}}
    \caption{Phase diagram for \(N{=}50\), \(T{=}2000\) across the \((p,q)\) plane.}
    \label{fig:pred_phase_diagr_50x2000}
\end{figure*}

\begin{figure*}[!t]
    \centering
    \subfloat[ROC-AUC]{\includegraphics[width=0.31\textwidth]{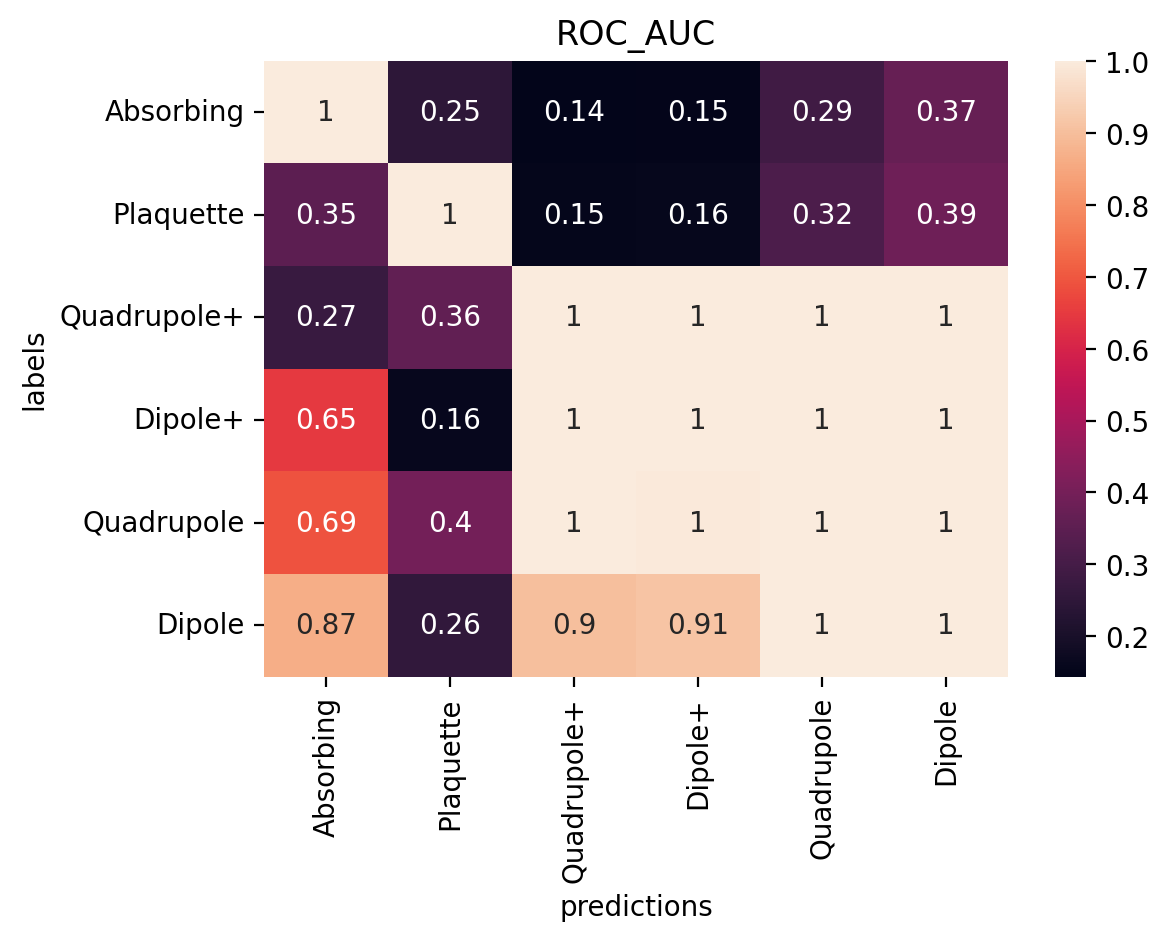}}\hfil
    \subfloat[PR-AUC]{\includegraphics[width=0.31\textwidth]{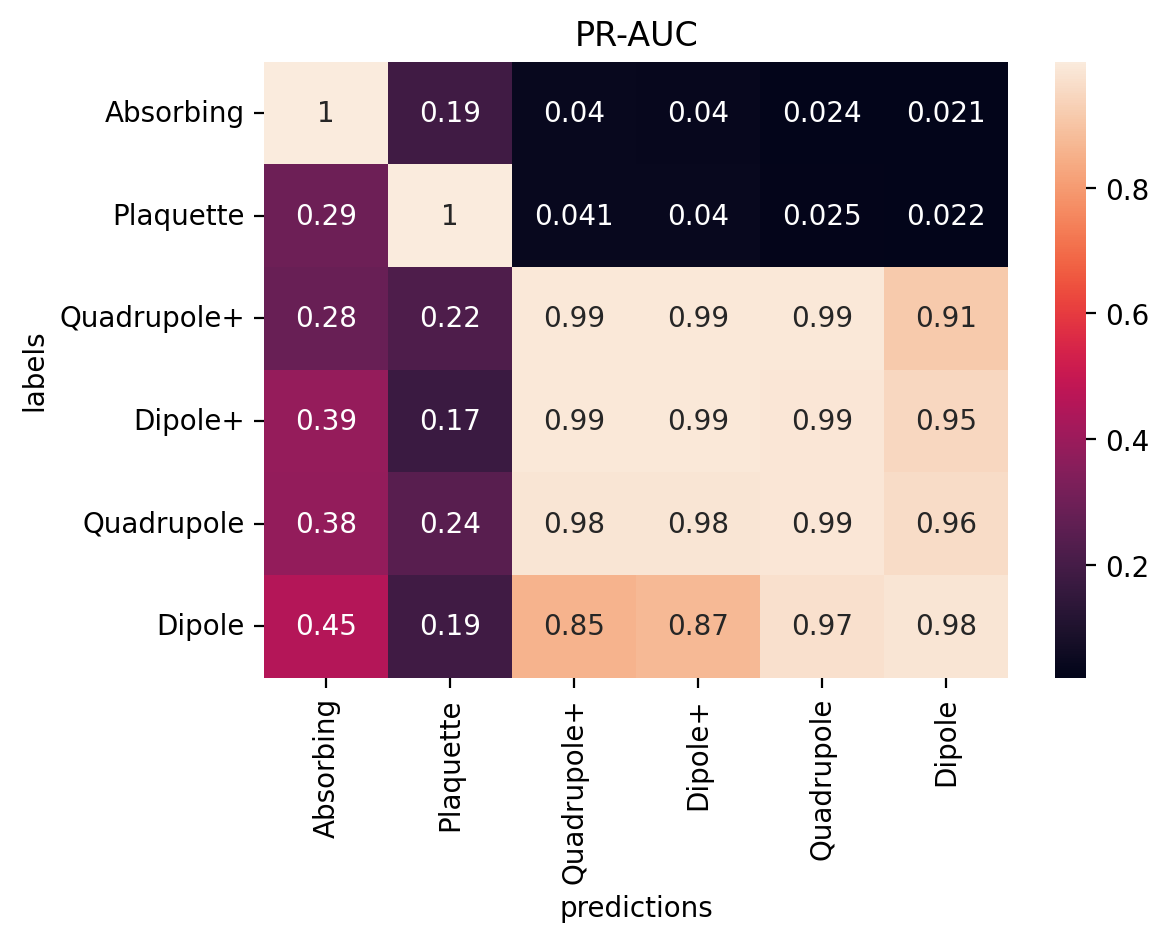}}\hfil
    \subfloat[$F_1$]{\includegraphics[width=0.31\textwidth]{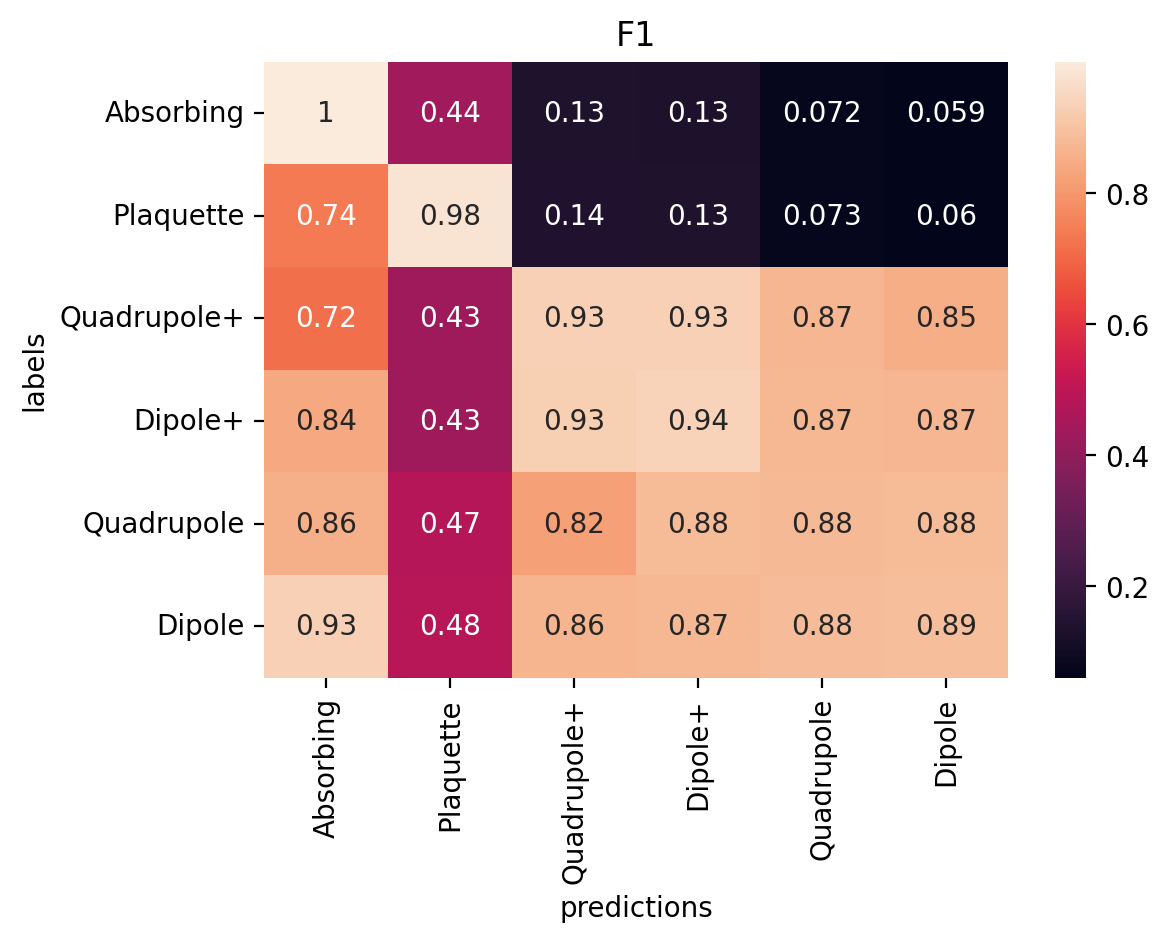}}
    \caption{ROC-AUC, PR-AUC, $F_1$ heatmaps of the form “metric (score of head $k$ \textrightarrow\ ground-truth class $c$)”.}
    \label{fig:metrics-heatmaps}
\end{figure*}

We also visualize matrices of the form “metric (score of head $k$ \textrightarrow\ ground-truth class $c$)” (Fig. ~\ref{fig:metrics-heatmaps}). ROC–AUC heatmaps reveal that, e.g., the $D^{+}$ head ranks not only $D^{+}$ well but also $D,Q,Q^{+}$ (positively), while anti-correlating with $PL$ and $A$ (high ROC-AUC for inverted labels). The PR-AUC heatmaps also confirm this, showing that the result is independent of the choice of threshold. On $F_1$ heatmaps (with best $\tau_k$) this effect is attenuated by thresholding but the hierarchy persists. Practical takeaway is that some heads can serve as informative proxies for other targets.

We report (i) per-head ROC-AUC and PR-AUC with bootstrap confidence intervals (resampling systems); (ii) per-head $F_1$ at the selected thresholds $\tau_k$; (iii) the cross-class ROC-AUC/PR-AUC/$F_1$ heatmaps; and (iv) calibration curves and reliability plots for post-sigmoid probabilities.

\section{Numerical experiments}\label{sec:experiments}

\subsection{Phase-diagram reproduction}\label{sec:phase-diagr-reproduction}
Using the calibrated decision rule from Sec.~\ref{sec:calibration}, the model reproduces the target phase diagram on dense \((p,q)\) grids. Figure~\ref{fig:pred_phase_diagr_50x2000} shows a representative result of a model trained on systems of size \(N{=}50\), \(T{=}1000\) and evaluated on systems of size \(N{=}50\), \(T{=}2000\).

\subsection{Subset-of-Heads Calibrator}\label{sec:phase-diagr-reproduction-by-subset}
Using a subset of heads ($A$, $Q$, $Q^{+}$, $PL$), the lightweight calibrator (Sec.~\ref{sec:calibration}) achieves phase diagram reconstruction of comparable quality and acceptable critical point deviation (Table~\ref{CritPoints}). This confirms that the network captures global features and reduces reliance on the full set of heads for single-label mapping.

\subsection{Bernoulli distribution}\label{sec:bernoulli}
To test whether the network hallucinates structure where none should exist, we performed a negative control on undirected (isotropic) configurations: for each \(p_b\in[0,1]\) we generated 1024 \emph{i.i.d.} Bernoulli fields of size \(N=50\), \(T=2000\) (each cell active with probability \(p_b\)), with no dynamics or directionality, then passed them through the model and averaged per-head probabilities. 

First, for no value of \(p_b\) does the network detect exclusive dipolar/quadrupolar patterns \(D, D^{+}, Q, Q^{+}\) --- consistent with these structures arising from the directed PCA dynamics and being absent in isotropic random fields. Second, at high fill fractions we do observe an isotropic plaquette (PL) signal, as large \(2{\times}2\) blocks appear purely combinatorially. Finally, a nonzero percolation score appears already at relatively small \(p_b\); this is an artifact/hallucination of the classifier, which interprets random late-time activations as a directed spanning cluster. 

In genuine PCA data, percolation cannot occur without continuous temporal ``ridges''; such ridges are absent in the isotropic controls, and the network indirectly confirms this by not inventing \(D/D^{+}/Q/Q^{+}\). This control highlights the importance of calibration and negative tests when interpreting percolation probabilities.

\subsection{Small-system stochasticity and length robustness}\label{sec:small-system}
\begin{figure}[!t]
    \centering
    \includegraphics[width=\linewidth]{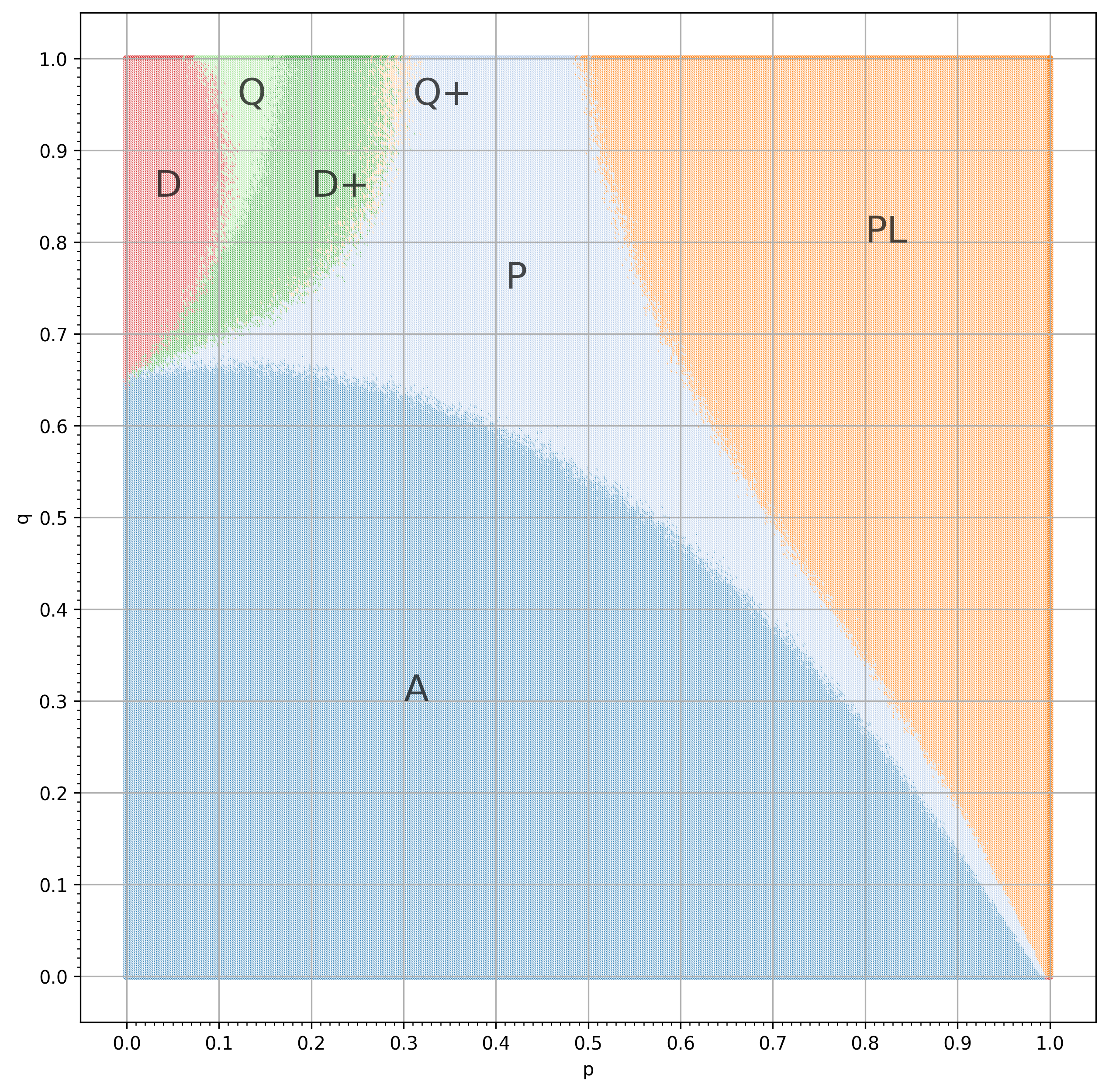}
    \caption{Target phase diagram for \(N{=}50\), \(T{=}1000\) across the \((p,q)\) plane.}
    \label{fig:fact_50x1000}
\end{figure}

For small systems (e.g., \(50\times 500\)) boundary stochasticity leads to noisy raw labels near transitions, as illustrated in Figure~\ref{fig:fact_50x1000}. After calibration (Sec.~\ref{sec:calibration}) the predicted maps become crisper; increasing \(T\) further sharpens boundaries (see Fig. ~\ref{fig:phase-multiT}), consistent with genuine temporal-dependency learning.

\subsection{Estimating critical points from calibrated probability sweeps}\label{sec:critical}
 
\begin{figure}[H]
    \centering
    \includegraphics[width=0.9\linewidth]{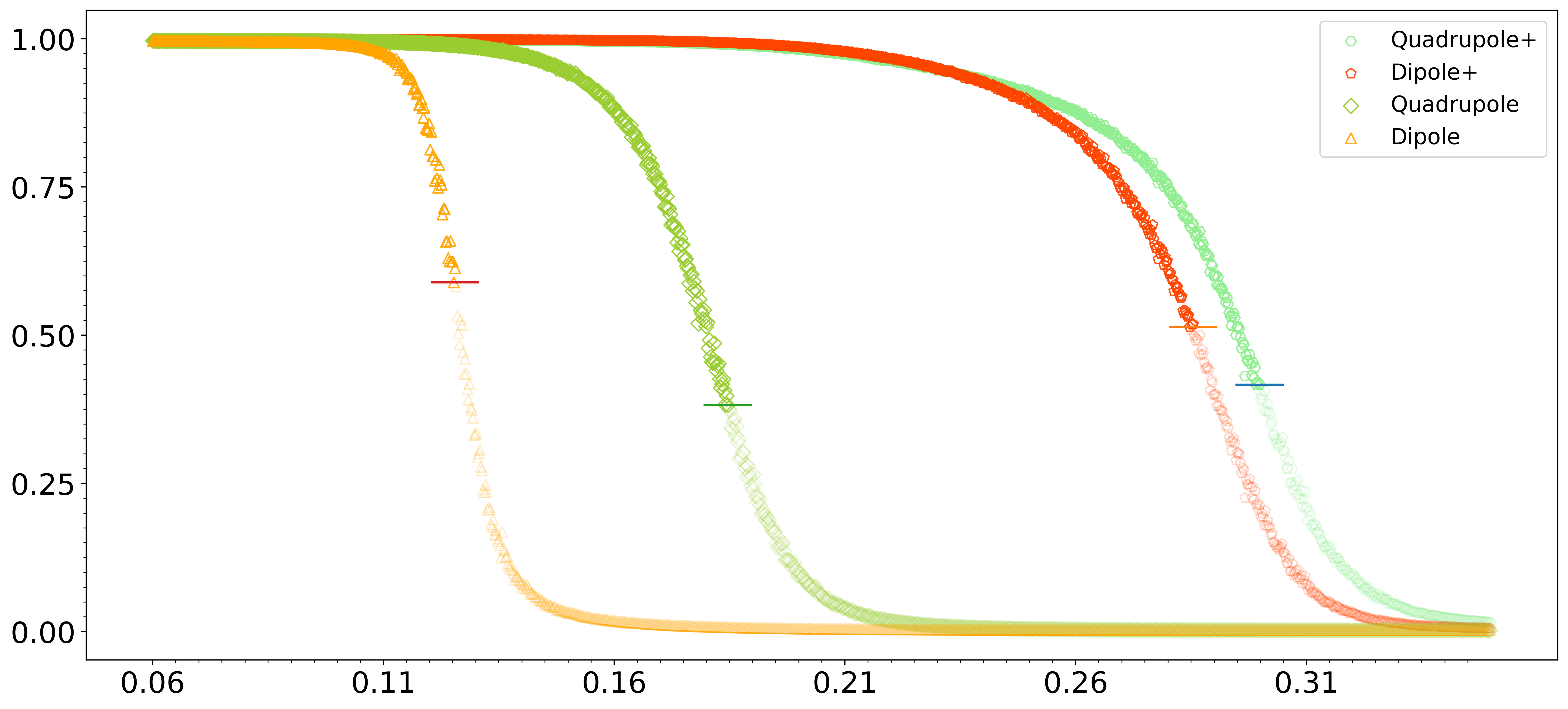}
    \caption{Calibrated probability sweep of the Dipole, Dipole+, Quadrupole, Quadrupole+ phases vs.\ $p$ in the range $[0.06,\,0.35]$ at fixed $q{=}0.9$.}
    \label{fig:targetphasetransition-a}
\end{figure}

\begin{figure}[H]
    \centering
    \includegraphics[width=0.9\linewidth]{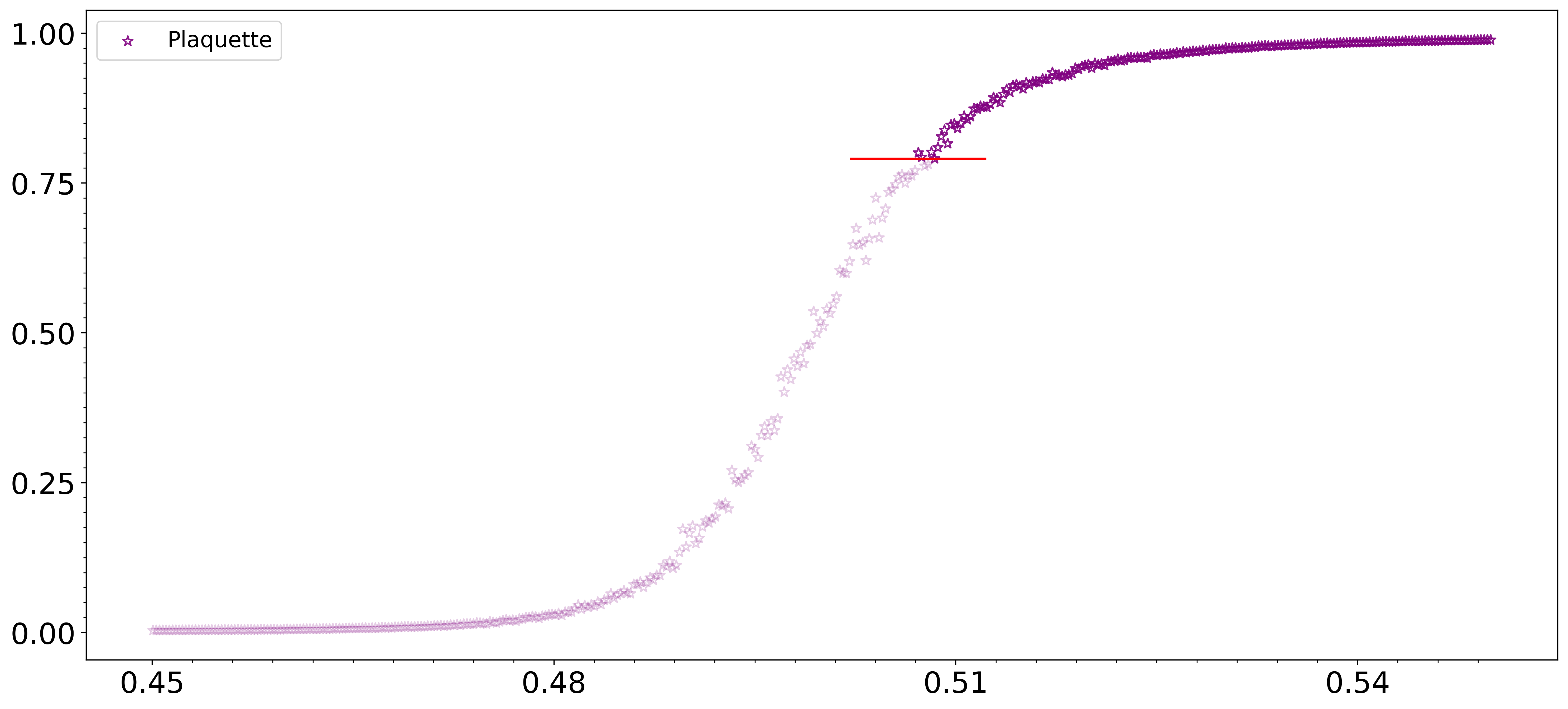}
    \caption{Calibrated probability sweep of the plaquette phase vs.\ $p$ in the range $[0.45,\,0.55]$ at fixed $q{=}0.9$.}
    \label{fig:targetphasetransition-b}
\end{figure}

We obtain indirect critical-point estimates by sweeping \(p\) at fixed \(q\) and detecting threshold crossings of calibrated per-head probabilities, see Fig.~\ref{fig:targetphasetransition-a}, ~\ref{fig:targetphasetransition-b}:
\begin{enumerate}
    \item Fix \(q=0.9\).
    \item For each \(p\) on a fine grid score 1024 configurations. We use two windows, \(p\in[0.06,0.35]\) and \(p\in[0.75,0.85]\), covering all observed transitions.
    \item Plot \(p\) (x-axis) vs.\ calibrated probability for a given pattern head (y-axis); sub-threshold points are shown semi-transparent, super-threshold points solid.
    \item The crossing yields the critical \(p\) for that pattern. For example, we observe exit from \(Q\) near \(p\approx 0.18\) at \(q=0.9\) (destination phase is not specified at this step).

\begin{table}[H]
    \centering
    \caption{Critical points at $q=0.9$}
    \begin{tabular}{|c|c|c|c|c|c|}
        \hline
        pattern & $Q^{+}$ & $D^{+}$ & $Q$ & $D$ & $PL$ \\
        \hline
        $p_0$ & $0.330$ & $0.315$ & $0.165$ & $0.120$ & $0.510$ \\
        \hline
        $p_1$ & $0.300$ & $0.285$ & $0.185$ & $0.126$ & $0.507$ \\
        \hline
        $p_2$ & $0.298$ & $0.293$ & $0.168$ & $0.146$ & $0.550$ \\
        \hline
        $p_3$ & $0.309$ & $0.305$ & $0.171$ & $0.112$ & $0.508$ \\
        \hline
    \end{tabular}
    \label{CritPoints}
\end{table}
\end{enumerate}

The numerical results are summarized in Table~\ref{CritPoints}, where $p_0$ represents the baseline critical points from the phase diagram of the previous work \cite{ovchinnikov2025hidden}, against which we compare the points obtained from the neural network inference: $p_1$ for systems of size 50 × 2000; $p_2$ for systems of size 50 × 2000 in the case of the calibrator applied to a subset of heads ($A$, $Q$, $Q^{+}$, and $PL$); and $p_3$ for systems of size 100 × 3000.

\section{Conclusion}\label{sec:conclusion}

We trained a multi-head network on raw space--time binary configurations to classify \([A,PL,Q^{+},D^{+},Q,D]\) phases of the replication model and, using PCA-based supervision, reproduced the phase diagram without handcrafted features. This confirms the earlier results of the direct numerical and scaling analyses \cite{Chitov:2015,Ovchinnikov:2025} that the internal percolation patterns are present in the raw data and can be detected directly.

A simple calibration pipeline (either a shallow auxiliary mapper or per-head PR thresholds with a fixed physical order) yields clean phase maps and remains robust across sequence lengths; larger \(T\) produces sharper boundaries.

Inference speed enables dense \((p,q)\) sweeps; preliminary tests show a clear advantage over deterministic pattern extraction (timings to be reported). Finally, calibrated probability sweeps versus \(p\) at fixed \(q\) provide an efficient indirect route to estimate critical points for individual patterns (e.g., exit from \(Q\) near \(p\approx 0.18\) at \(q=0.9\)).

\textit{Limitations and outlook.} Class imbalance and boundary stochasticity make thresholds sensitive to sampling; future work will integrate improved uncertainty calibration, balanced samplers, semi-supervised targets beyond PCA, and larger \((N,T)\) to refine critical-point estimates and to search for additional hidden structures.These results suggest that neural networks can be successfully trained for the search of complex connectivity patterns in large model-generated or empirical connected graphs (networks) in future work. 

\section*{Acknowledgment}
This research was financially supported by a grant $\# 24-22-00075$
(\href{https://rscf.ru/en/project/24-22-00075/}{https://rscf.ru/en/project/24-22-00075/}) from the Russian Science Foundation.
VK acknowledges the research environment provided by RCE I-FIM (MoE, Singapore). 

\bibliographystyle{unsrt}
\bibliography{ARTICLE_NN.bib}

\end{document}